\documentclass[10pt, a4paper]{article}
\usepackage{lrec}
\usepackage{graphicx}
\usepackage{multirow}
\usepackage[english]{babel}
\usepackage{CJKutf8}

\usepackage{epstopdf}

\usepackage{hyperref}

\title{A Discourse-Level Named Entity Recognition and Relation Extraction Dataset \\for Chinese Literature Text}

\name{Jingjing Xu\footnotemark{*}, Ji Wen\footnotemark{*}\thanks{* Equal Contribution.}, Xu Sun, Qi Su}

\address{MOE Key Laboratory of Computational Linguistics, Peking University\\
School of Electronics Engineering and Computer Science, Peking University\\
\{jingjingxu,wenjics,xusun,sukia\}@pku.edu.cn}

\abstract{
Named Entity Recognition and Relation Extraction for Chinese literature text is regarded as the highly difficult problem, partially because of the lack of tagging sets. In this paper, we build a discourse-level dataset from hundreds of Chinese literature articles for improving this task. To build a high quality dataset, we propose two tagging methods to solve the problem of data inconsistency, including a heuristic tagging method and a machine auxiliary tagging method. Based on this corpus, we also introduce several widely used models to conduct experiments. Experimental results not only show the usefulness of the proposed dataset, but also provide baselines for further research. 
The dataset is available at https://github.com/lancopku/Chinese-Literature-NER-RE-Dataset.
\newline  }

\begin{document}

\maketitleabstract

\section{Introduction}

Recent researches on Named Entity Recognition (NER)~\cite{LinW09,collobert2011natural,HuangXY15} and Relation Extraction (RE)~\cite{Kambhatla2004,zeng2014relation,nguyen2015combining} focused on news articles and achieved the promising performance. However, for a complex but important work, Chinese literature, this task becomes more difficult due to the lack of datasets. Thus, in this paper, we build a NER and RE dataset from hundreds of Chinese literature articles. Unlike previous sentence-level datasets where sentences are independent with each other, we build a discourse-level dataset where sentences from the same passage provide the additional context information. 

\begin{figure}
  \centerline{\includegraphics[width=0.8\linewidth]{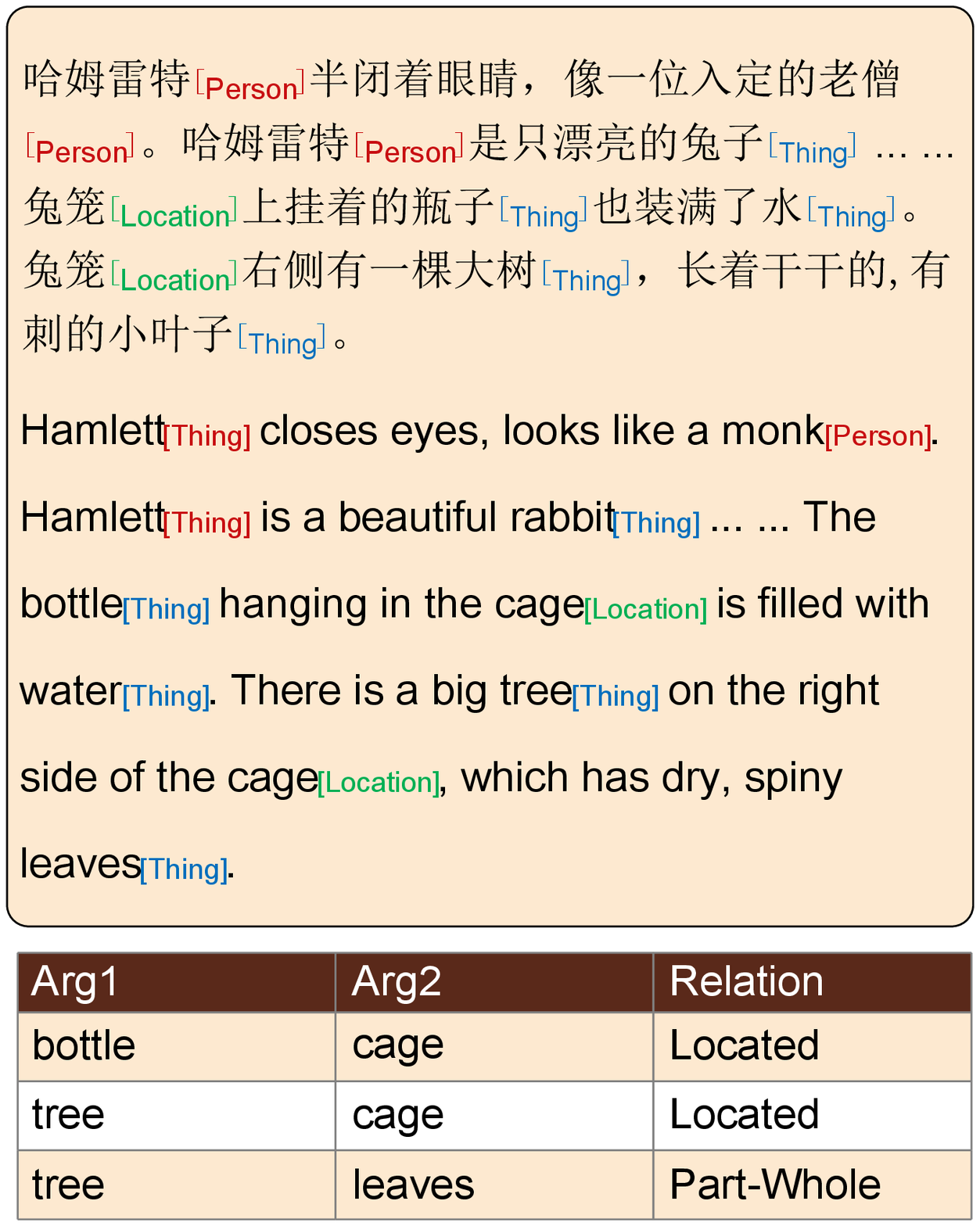}}
\caption{A tagging example. The top table describes the raw text and tagged entities which are shown in different color. The bottom table shows the tagged relations among these entities. }
\label{example}
\end{figure}

However, tagging entities and relations in Chinese literature text is more difficult than that in traditional datasets which have simple entity classes and explicit relationships. Various rhetorical devices  pose great challenges for building a high-consistency dataset. A simple example of personification is shown in Figure~\ref{example}. ``Hamlett'' is a person name but refers to a rabbit. Some annotators label it with a ``Person'' tag and another annotators label it with a ``Thing'' tag. Thus, the major difficulty lies in how to handle massive ambiguous cases to ensure data consistency.


In this paper, we propose two methods to solve this problem. On one hand, we define several generic disambiguating rules to deal with the most common cases. On the other hand, since these heuristic rules are too generic to handle all ambiguous cases, we also introduce a machine auxiliary tagging method which uses the annotation standards learned from the subset of the corpus to predict labels on the rest data. Annotators just care about the cases where the predicted labels are different with the gold labels, which significantly reduces annotators' efforts. 

Overall speaking, we manually annotate 726 articles, 29,096 sentences and over 100,000 characters in total, which is accomplished in 300 person-hours spread across 5 people and three months. 

\begin{figure*}[!hbt]
  \centerline{\includegraphics[width=0.9\textwidth]{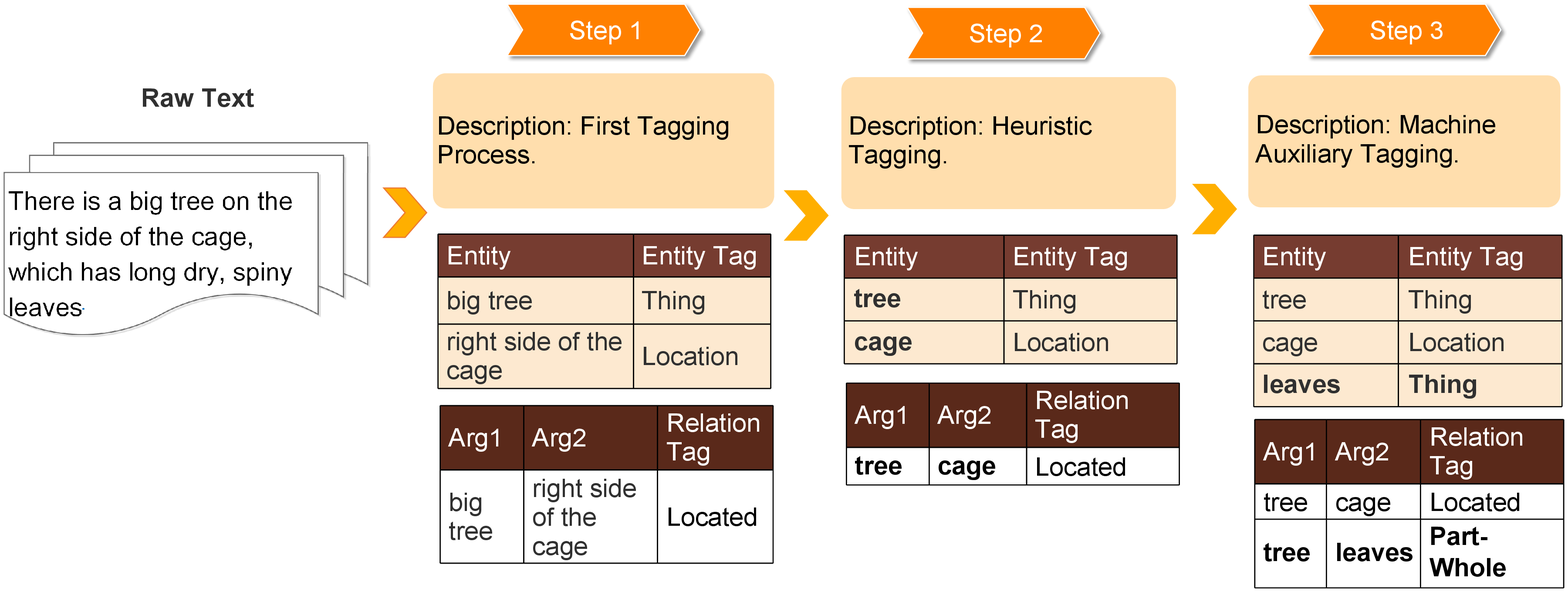}}
\caption{Illustration of the annotation process.}
\label{dsns}
\end{figure*}

Based on this corpus, we also introduce some widely used models to conduct experiments. Experimental results not only show the usefulness of the proposed dataset, but also provide baselines for further research. 

 Our contributions are listed as follows:
\begin{itemize}
\item We provide a new dataset for joint learning of Named Entity Recognition and Relation Extraction for Chinese literature text.

\item Unlike previous sentence-level datasets, the proposed dataset is based on the discourse level which provides additional context information.


%

\item Based on this corpus, we introduce some widely used models to conduct experiments which can be used as baselines for further works.
\end{itemize}

\begin{CJK*}{UTF8}{gbsn}
\begin{table*}[!hbt]
\centering

\begin{tabular}{cccc}
\hline
Tags&Descriptions&Examples&\%\\
\hline
Thing& Thing& 苹果 (apple)&35.63\\

\hline
Person& Person& 李秋 (Qiu Li)&32.71\\
\hline
Location& Location, country or city name& 巴黎 (France)&17.18\\

\hline
Time& Time related words& 一天 (one day)&7.36\\
\hline
Metric& Measurement related words&一升 (1L)&3.64
\\
\hline
Organization& Organization name& 信息学报 (Journal of Information Processing)&2.03\\
\hline
Abstract& Abstract & 山西日报 (Shanxi daily)&1.43\\
\hline
\end{tabular}
\caption{ The set of entity tags. The
last column indicates each tag's relative frequency in the
full annotated data. }
\label{entitytags}
\end{table*}
\end{CJK*}

\begin{CJK*}{UTF8}{gbsn}
\begin{table*}[!hbt]
\centering

\begin{tabular}{cccc}
\hline
Tags&Descriptions&Examples&\%\\
\hline
Located&Be located in&  幽兰 (orchid)-山谷 (valley)&37.43\\

\hline
Part-Whole&Be a part of&  花 (flower)-仙人掌 (cactus)&23.76\\
\hline
Family&Family relationship&  母亲 (mother)-奶奶 (grandmother)&10.25\\

\hline
General-Special&Generalization relationship&  鱼 (fish)-鲫鱼 (carp)&6.99\\

\hline
Social&Be socially related&  母亲 (mother)-邻里 (neighbour)&6.02\\

\hline
Ownership&Occupation relationship&  村民 (villager)-旧屋 (house)&5.10\\

\hline
Use&Do something with&  爷爷 (grandfather)-毛笔 (brush)&4.76\\

\hline
Create&Bring about something&  男人 (man)-陶器 (pottery)&2.93\\

\hline
Near&A short distance away&  山 (hill)-县城 (town)&2.76\\

\hline
\end{tabular}
\caption{ The set of relation tags. The
last column indicates each tag's relative frequency in the
full annotated data. }
\label{relation}
\end{table*}
\end{CJK*}

\section{Data Collections}

We first obtain over 1,000 Chinese literature articles from the website and then filter, extract 726 articles. Too short or too noise articles are not included. Due to the difficulty of tagging Chinese literature text, we divide the annotation process into three steps. The detailed tagging process are shown in Figure~\ref{dsns}

\textbf{Step 1: First Tagging Process.} We first attempt to annotate the raw articles based on defined entity and relation tags. In the tagging process, we find a problem of data inconsistency. To solve this problem, we design the next two steps.

\textbf{Step 2: Heuristic Tagging Based on Generic disambiguating Rules.}
We design several generic disambiguation rules to ensure the consistency of annotation guidelines. For example, remove all adjective words and only tag ``entity header'' when tagging entities (e.g., change ``a girl in red cloth'' to ``girl''). In this stage, we re-annotate all articles and correct all inconsistency entities and relations based on the heuristic rules. 


\textbf{Step 3: Machine Auxiliary Tagging.} Even though the heuristic tagging process significantly improves dataset quality, it is too hard to handle all inconsistency cases based on limited heuristic rules. Thus, we introduce a machine auxiliary tagging method. The core idea is to train a model to learn annotation guidelines on the subset of the corpus and produce predicted tags on the rest data. The predicted tags are used to be compared with the gold tags to discovery  inconsistent entities and relations, which largely reduce annotators' efforts. Specifically, we divide the corpus into 10 parts, and make predictions on each part of the corpus based on the model trained on the rest of the corpus. The model we used in this paper is CRF with a simple bigram feature template. 

After all annotation steps, we also check all entities and relations to ensure the correctness of the dataset.

\section{Data Properties}
We will describe the tagging set and annotation format in detail.

\subsection{Tagging Set}

We define 7 entity tags and 9 relation tags based on several available NER and RE datasets but with some additional categories specific to Chinese literature text. Details of the tags are shown in Table~\ref{entitytags} and \ref{relation}.

We add three new entity tags specific for understanding literature text, including ``Thing'', ``Time'' and ``Metric''. ``Thing'' is for capturing  objects which articles mainly describe, such as ``flower'', ``tree'' and so on. ``Time'' is  for capturing the time-line of a story, such as ``one day'', ``one month'' and so on. ``Metric'' is for capturing the measurement related words, such as ``1L'', ``1mm'' and so on.

As for relation tags, we set 9 different classes for better understanding the connection between entities, including ``Located'', ``Near'', ``Part-Whole'', ``Family'', ``Social'', ``Create'', ``Use'', ``Ownership'', ``General-Special''. For building the relations between people in literature articles, we use the ``Social'' tag, which is not quite common in other corpora.

\subsection{Annotation Format}

Each entity is identified by ``T'' tag, which takes several attributes.

\begin{itemize}
\item \textsl{Id}: a unique number identifying the entity within
 the document. It starts at 0, and is incremented every
time a new entity is identified within the same
document. 

\item \textsl{Type}: one of the entity tags.
 
\item \textsl{Begin Index}: the begin index of an entity. It starts at 0, and is incremented every character.

\item \textsl{End Index}: the end index of an entity. It starts at 0, and is incremented every character.

\item \textsl{Value}: words being referred to an identifiable object.
\end{itemize}

Each relation is identified by ``R'' tag, which can take several attributes: 

\begin{itemize}
\item \textsl{Id}: a unique number identifying the relation within
the document. It starts at 0, and is incremented every
time a new relation is identified within the same
document. 

\item \textsl{Arg1 and Arg2}: two entities associated with a relation.

\item \textsl{Type}: one of the relation tags.
\end{itemize}

\begin{table*}
\centering
\begin{tabular}{c|c|ccccccc}
\hline
\multicolumn{1}{c|}{\multirow{1}{*}{Models}}&&\multicolumn{1}{c}{\multirow{1}{*}{Thing}}&\multicolumn{1}{c}{\multirow{1}{*}{Person}}&\multicolumn{1}{c}{\multirow{1}{*}{Location}}&\multicolumn{1}{c}{\multirow{1}{*}{Organization}}&\multicolumn{1}{c}{\multirow{1}{*}{Time}}&\multicolumn{1}{c}{\multirow{1}{*}{Metric}}&\multicolumn{1}{c}{\multirow{1}{*}{All}}\\
\hline

\multicolumn{1}{c|}{\multirow{3}{*}{Bi-LSTM}}
&P&67.07&80.30&58.09&Nan&64.47&46.15&70.52\\
&R&62.37&78.50&46.79&Nan&45.51&22.18&62.36\\
&F&64.63&79.39&51.83&Nan&53.36&29.96&66.19\\
\hline


\multicolumn{1}{c|}{\multirow{3}{*}{CRF}}
&P&75.72&87.92&68.41&46.69&76.20&70.50&77.72\\
&R&65.42&82.27&50.98&45.26&60.93&38.42&65.91\\
&F&70.19&85.00&58.42&45.96&67.72&49.74&71.33\\
\hline
\end{tabular}

\caption{Results of Named Entity Recognition on the proposed corpus.
}\label{daimprovements}
\end{table*}

\begin{table*}
\centering
\begin{tabular}{c|c|c}
\hline
\multicolumn{1}{c|}{\multirow{1}{*}{Models}}&Information&\multicolumn{1}{c}{\multirow{1}{*}{ $F_{1}$ }}\\

\hline
SVM&Word embeddings, NER, WordNet, HowNet,  &\multicolumn{1}{c}{\multirow{2}{*}{48.9}}\\
\cite{Hendrickx2010}&POS, dependency parse, Google n-gram&\\

\hline

RNN&Word embeddings&48.3\\
\cite{socher2011semi}&+ POS, NER, WordNet&49.1\\

\hline

CNN&Word embeddings&47.6\\
\cite{zeng2014relation}&+ word position embeddings, NER, WordNet&52.4\\

\hline

CR-CNN&Word embeddings&52.7\\
\cite{santos2015classifying}&+ word position embeddings&54.1\\

\hline

SDP-LSTM&Word embeddings&54.9\\
\cite{xu2015classifying}&+ POS + NER + WordNet&55.3\\

\hline

DepNN&\multicolumn{1}{c|}{\multirow{2}{*}{Word embeddings, WordNet}}&\multicolumn{1}{c}{\multirow{2}{*}{55.2}}\\
\cite{LinW09}&&\\
\hline

BRCNN&Word embeddings&55.0\\
\cite{cai2016bidirectional}&+ POS, NER, WordNet&55.6\\





\hline
\end{tabular}

\caption{Results of Relation Extraction on the proposed corpus.}\label{relations}
\end{table*}

\section{Experiments}
We introduce several baselines to conduct experiments. In this section we will describe experiment settings, baselines and experiment results in detail.

\subsection{Settings}
 Experiments are performed on a commodity 64-bit Dell Precision T5810 workstation with one 3.0 GHz 16-core CPU and 64GB RAM. The performance of NER and RE models are evaluated by $F_{1}$-score. For training, we use mini-batch stochastic gradient descent to minimize negative log-likelihood. Training is performed with shuffled mini-batches of size 32.

\subsection{Named Entity Recognition} 
We introduce LSTM~\cite{hochreiter1997long} and CRF~\cite{Lafferty01conditionalrandom} as our baselines, which are described as follows.

\textbf{LSTM}
We consider bi-directional LSTM as one of models. Both the character embedding dimension and the hidden dimension are set to be 100. 


\textbf{CRF} CRF is a statistical modeling method which often applied in pattern recognition and machine learning. Our features template includes unigram and bigram features.

The results are shown in Table~\ref{daimprovements}. It can be clearly seen that CRF achieves the better performance than Bi-LSTM on all tags, which probably be attributed to the feature template. 

All models perform better on ``Person'', ``Thing'' and ``Time'' tags than on ``Location'', ``Organization'' and ``Metric'' tags. It shows that ``Person'', ``Thing'' and ``Time'' tags are the most easily identifiable entities. The problem of data sparsity makes it hard for capturing ``Location'', ``Organization'' and ``Metric'' tags. 

The higher accuracies show that the entities predicted by the model probably are the right tags, which reflects the data consistency between the training and testing sets. The lower recalls indicate that there is still a lot of unknown entities on the testing set. How to handle these unknown entities is a urgent problem for further research.

\subsection{Relation Extraction}
Table~\ref{relations} compares several state-of-the-art methods on the proposed corpus. The first entry in the table presents the highest performance achieved by traditional feature-based methods. Hendrickx et al.~\shortcite{Hendrickx2010} feeds a variety of hand-crafted features to the SVM classifier and achieves an $F_{1}$-score of 48.9.

Recent performance improvements on the task of relation classification are mostly achieved with the help of neural networks. Socher et al.~\shortcite{socher2011semi} builds a recursive neural network on the constituency tree and achieves a comparable performance with Hendrickx et al.~\shortcite{Hendrickx2010}.  Xu et al.~\shortcite{xu2015classifying} introduces a type of gated recurrent neural network which could raise the $F_{1}$ score to 55.3. By diminishing the impact of the other classes, Santos et al.~\shortcite{santos2015classifying} achieves an $F_{1}$-score of 54.1. Along the line of CNNs, Liu et al.~\shortcite{LinW09} achieves an $F_{1}$-score of 55.2.


\section{Conclusions}
We build a discourse-level Named Entity Recognition and Relation Extraction dataset for Chinese literature text. To solve the problem of data inconsistency in tagging process, we propose two methods in this paper, one is a heuristic tagging method and another is a machine auxiliary tagging method. Based on this corpus, we introduce several widely-used models to conduct experiments, which provides baselines for further research.

\nocite{DBLP:journals/corr/MaS17}

\section{References}

\bibliographystyle{lrec}
\bibliography{xample}

\end{document}